\documentclass[10pt, conference, compsocconf]{IEEEtran}
\ifCLASSINFOpdf
\else
\fi
\usepackage{graphicx}
\usepackage{subfigure}
\usepackage{amsmath}
\usepackage{multirow}
\graphicspath{{figures/}}
\makeatletter
\newif\if@restonecol
\makeatother

\usepackage[linesnumbered,boxed,ruled]{algorithm2e}
\DeclareMathOperator*{\argmin}{argmin}
\DeclareMathOperator*{\argmax}{argmax}

\begin{document}
%
\title{On the Feature Discovery for App Usage Prediction in Smartphones}


\author{Submitted for Blind Review}

\author{\IEEEauthorblockN{Zhung-Xun Liao\IEEEauthorrefmark{1},
Shou-Chun Li\IEEEauthorrefmark{1},
Wen-Chih Peng\IEEEauthorrefmark{1} and
Philip S Yu\IEEEauthorrefmark{2}}
\IEEEauthorblockA{\IEEEauthorrefmark{1}Department of Computer Science,\\
National Chiao Tung University, HsinChu, Taiwan \\
Email: \{zxliao.cs96g, scli.cs02g, wcpeng\}@nctu.edu.tw}
\IEEEauthorblockA{\IEEEauthorrefmark{2}Department of Computer Science,\\
University of Illinois at Chicago, Chicago, IL, USA \\
Email: psyu@cs.uic.edu}}


%


\maketitle
\begin{abstract}
With the increasing number of mobile Apps developed, they are now closely integrated into daily life. In this paper, we develop a framework to predict mobile Apps that are most likely to be used regarding the current device status of a smartphone. Such an Apps usage prediction framework is a crucial prerequisite for fast App launching, intelligent user experience, and power management of smartphones. By analyzing real App usage log data, we discover two kinds of features: The Explicit Feature (EF) from sensing readings of built-in sensors, and the Implicit Feature (IF) from App usage relations. The IF feature is derived by constructing the proposed App Usage Graph (abbreviated as AUG) that models App usage transitions. In light of AUG, we are able to discover usage relations among Apps. Since users may have different usage behaviors on their smartphones, we further propose one personalized feature selection algorithm. We explore minimum description length (MDL) from the training data and select those features which need less length to describe the training data. The personalized feature selection can successfully reduce the log size and the prediction time. Finally, we adopt the kNN classification model to predict Apps usage. Note that through the features selected by the proposed personalized feature selection algorithm, we only need to keep these features, which in turn reduces the prediction time and avoids the curse of dimensionality when using the kNN classifier. We conduct a comprehensive experimental study based on a real mobile App usage dataset. The results demonstrate the effectiveness of the proposed framework and show the predictive capability for App usage prediction.
\end{abstract}

\begin{IEEEkeywords}
Mobile Application; Usage Prediction; Classification; Apps;

\end{IEEEkeywords}

%
\IEEEpeerreviewmaketitle

\section{Introduction}
With the increasing number of smartphones, mobile applications (Apps) have been developed rapidly to satisfy users' needs~\cite{DBLP:conf/wsdm/YinLLW13,DBLP:conf/icmi/DoBG11,getjar,appjoy}. Users can easily download and install Apps on their smartphones to facilitate their daily lives. For example, users use their smartphones for Web browsing, shopping and socializing~\cite{DBLP:journals/tkde/LuLT12,DBLP:conf/huc/ChoujaaD10}. By analyzing the collected real Apps usage log data, the average number of Apps in a user's smartphone is around 56. For some users, the number of Apps is up to 150. As many Apps are installed on a smartphone, users need to spend more time swiping screens and finding the Apps they want to use. From our observation, each user has on average 40 launches per day. In addition, the launch delay of Apps becomes longer as their functionality becomes more complicated. In~\cite{DBLP:conf/mobisys/YanCGKL12}, the authors investigated the launch delay of Apps. Even simple Apps (e.g., weather report) need 10 seconds, while complicated Apps (e.g., games) need more than 20 seconds to reach a playable state. Although some Apps could load stale content first and fetch new data simultaneously, they still need several seconds to complete loading. 

To ease the inconvenience of searching for Apps~\cite{DBLP:conf/icdm/LiaoLSLP12,DBLP:conf/huc/ShinHD12} and to reduce the delay in launching Apps~\cite{DBLP:conf/mobisys/YanCGKL12}, one possible way is to predict which Apps will be used before the user actually needs them. Although both the iOS and Android systems list the most recently used (MRU) Apps to help users relaunch Apps, this method only works for those Apps which would be immediately relaunched within a short period. Another common method is to predict the most frequently used (MFU) Apps. However, when a user has a lot of frequently used Apps, the MFU method has very poor accuracy. In our experiments, these two methods are the baseline methods for comparison. 

Recently, some research works have addressed the Apps usage prediction problems~\cite{DBLP:conf/mobisys/YanCGKL12,DBLP:conf/icdm/LiaoLSLP12,DBLP:conf/huc/ShinHD12}.  In~\cite{DBLP:conf/icdm/LiaoLSLP12}, a temporal profile is built to represent the usage history of an App. The temporal profile records the usage time and usage period of the App. Then, when a query time is given, the usage probability of each App could be calculated through comparing the difference between the temporal profile and the query time. However, since they only consider the periodicity feature of Apps, some Apps with no significant periods cannot be predicted by their temporal profiles. In~\cite{DBLP:conf/mobisys/YanCGKL12}, the authors adopted three features to predict Apps usage: time, location, and used Apps. Based on those three features, they designed and built a system to remedy slow App launches. However, they always use these three features to predict different users' usage, which is impractical as users could have different usage behavior. For example, the location information could be less useful for those users who have lower mobility. We claim that the features which are able to accurately predict Apps usage are different for different users and different Apps. The authors in~\cite{DBLP:conf/huc/ShinHD12} collected 37 features from accelerometer, Wi-Fi signal strength, battery level, etc., and proposed a Naive Bayes classification method to predict Apps usage. However, a Naive Bayes classification method needs sufficient training data to calculate the conditional probability, which does not always hold. Therefore, the system would fail to predict Apps if there are not exactly the same instances existing in the training dataset. In addition, they still apply all the same features to each user, instead of selecting personalized features for different users with different usage behaviors. 

\begin{figure}
	\centering
    \includegraphics[scale=0.45]{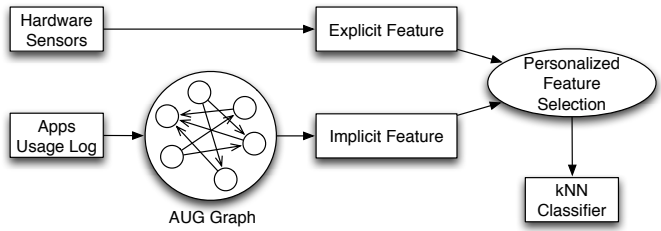}
    \caption {Overview of kNN-based App Prediction framework.}
	\label{fig:overview}
\end{figure}

In this paper, we adopt the concept of minimum description length (MDL) to select personalized features for different users and propose a kNN-based App Prediction framework, called KAP, to predict Apps usage. Once we distinguish the useful and useless features, only the useful features need to be collected. Therefore, the size of the log data could be reduced. The overall framework is shown in Figure~\ref{fig:overview}. KAP investigates features from both explicit and implicit aspects. The explicit feature is a set of sensor readings from built-in hardware sensors, such as GPS, time, accelerometers, etc. On the other hand, the implicit feature is referred to as the correlations of Apps usage. To capture these correlations, the implicit feature is represented as the transition probability among Apps.

For the explicit feature, we focus on three types of hardware sensors: 1) device sensors, such as free space, free ram, and battery level, 2) environmental sensors, such as time, GSM signal, and Wi-Fi signal, and 3) personal sensors: acceleration, speed, heading, and location. We claim that the usage of different Apps is related to different types of sensors. Obviously, the advantages of selecting sensors for the explicit feature is that it reduces the effect of noisy data and also saves power and storage consumption for logging data and performing the prediction.

For the implicit feature, we calculate the transition probability for each App. However, the previous works~\cite{DBLP:conf/mobisys/YanCGKL12,DBLP:conf/huc/ShinHD12} only take the usage order into account, and not the time duration between Apps. We claim that the length between Apps usage means different things. For example, users may take pictures via a camera App and upload those pictures to Facebook. However, some users may upload pictures immediately, while others would upload them when they have a Wi-Fi connection. Therefore, the time duration between camera and Facebook use depends on different users and different usage behaviors. To model the usage relation among Apps, an Apps Usage Graph (AUG), which is a weighted directed graph, is proposed. The weight on each edge is formulated as an exponential distribution to describe the historical usage durations. Based on AUG, the implicit feature of each training instance is derived by traversing the AUG. Consequently, the implicit feature of each testing case is derived by an iterative refinement process.   

With both explicit and implicit features, KAP adopts a kNN classification model to predict Apps usage which is represented as class labels. In the experimental study, the proposed KAP framework outperforms both baseline methods and achieves accuracy of 95\%. We also show that the personalized sensor selection for the explicit feature is efficient and effective. In addition, the implicit feature is useful for improving the prediction accuracy of KAP. 

The major contributions of this research work are summarized as follows.
\begin{itemize}

\item We address the problem of Apps usage prediction by discovering different feature sets to fulfill different users' Apps usage behavior, and propose the concept of explicit and implicit features for Apps usage prediction. 


\item We estimate the distribution of the transition probability among Apps and design an Apps Usage Graph (AUG) to model both Apps usage order and transition intervals. Two algorithms are proposed to extract the implicit features from the AUG graph for training and testing purposes respectively.


\item We propose a personalized feature selection algorithm in which one could explore MDL to determine a personalized set of features while still guaranteeing the accuracy of the predictions.

\item A comprehensive performance evaluation is conducted on real datasets, and our proposed framework outperforms the state-of-the-art methods~\cite{DBLP:conf/huc/ShinHD12}.
\end{itemize}

The rest of this paper is organized as follows. Section~\ref{sec:related} investigates the related works which discuss the conventional prediction problem and Apps usage prediction. Section~\ref{sec:feature} introduces the explicit and implicit features. Section~\ref{sec:selection} presents the mechanism of personalized feature selection. Section~\ref{sec:exp} conducts extensive and comprehensive experiments. Finally, this paper is concluded with Section~\ref{sec:conclusion}.
\section{related works}
\label{sec:related}
To the best of our knowledge, the prediction problem of Apps usage in this paper is quite different from the conventional works. We focus on not only analysing usage history to model users' behavior, but on personalizing varied types of features including hardware and software sensors attached to smartphones. The proposed algorithm selects different features for different users to satisfy their usage behavior. Although there have been many research works solving the prediction problem in different domains, such as music items or playlist prediction~\cite{DBLP:conf/kdd/ChenMTJ12}, dynamic preference prediction~\cite{pakdd13,DBLP:conf/sigir/LathiaHCA10}, location prediction~\cite{DBLP:conf/mdm/LeiSPS11,DBLP:conf/pervasive/ScellatoMMLC11,DBLP:conf/kdd/MonrealePTG09}, social links prediction~\cite{DBLP:conf/icdm/DongTWTCRC12,DBLP:conf/cikm/Liben-NowellK03}, and so on, the prediction methods are only based on analysing the usage history. In~\cite{DBLP:conf/sdm/Kogan12}, the author selected features from multiple data streams, but the goal is to solve the communication problem in a distributed system.

Currently, only a few studies discuss mobile Apps usage prediction. 
%
%
Although the authors in~\cite{inss08} adopted location and time information to improve the accuracy of Apps usage prediction, the total number of Apps is only 15. Concurrently, in~\cite{kam09}, the authors stated that the prediction accuracy could achieve 98.9\%, but they still only focus on predicting 9 Apps from a set of 15. In~\cite{DBLP:conf/mobisys/YanCGKL12}, the authors solved the prediction problem through multiple features from 1) location, 2) temporal burst, and 3) trigger/follower relation. However, they did not analyze the importance of each feature. Therefore, for different users, they always use the same three features to predict their Apps usage. In~\cite{DBLP:conf/huc/ShinHD12}, the authors investigated all possible sensors attached to a smartphone and adopted a Naive Bayes classification to predict the Apps usage. However, collecting all possible sensors is inefficient and impractical. Moreover, the useful sensors for different users could vary according to users' usage behavior. We claim that for different users, we need to use different sets of features to predict their usage. In this paper, we collect only the subset of all features which are personalized for different users.

This paper is the first research work which discusses how to select suitable sensors and features for different users to predict their Apps usage. Through the personalized feature selection, we could perform more accurate predictions for varied types of usage bahavior, reduce the dimensionality of the feature space, and further save energy and storage consumption. In addition, the proposed KAP framework derives the implicit feature by modelling the usage transition among Apps.
\section{Explicit and Implicit Features}
\label{sec:feature}
In this paper, we separate the features into two main categories: the explicit feature and the implicit feature. The explicit feature represents the sensor readings which are explicitly readable and observable. The implicit feature is the Apps usage relations.

\subsection{Explicit Feature Collection}
Table~\ref{tab:explicit} shows the hardware sensors we use for the explicit feature. As different models of smartphones could have different sets of hardware sensors, we only list the most common ones whose readings are easy to record. It is totally free to add or remove any hardware sensors here. 
 
\begin{figure}
	\centering
	\subfigure[$User_1$]{
    	\label{fig:sub:user1}
        \includegraphics[scale=0.8]{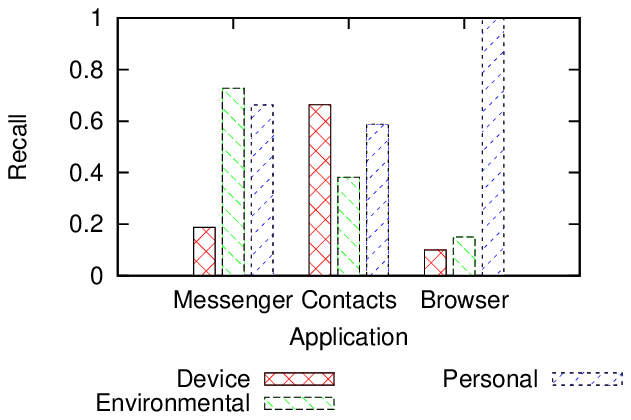}
    }
	\subfigure[$User_2$]{
    	\label{fig:sub:user2}
        \includegraphics[scale=0.8]{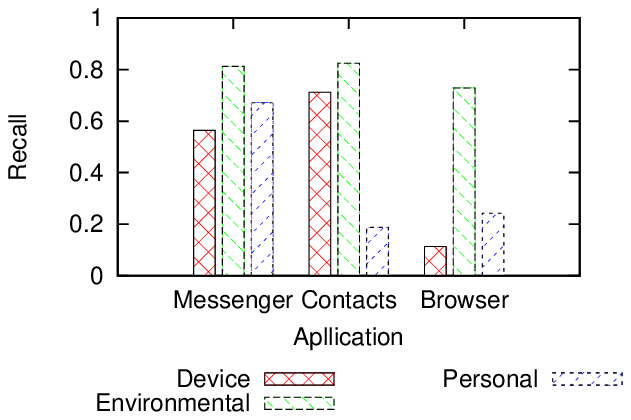}
    }
    \caption{Varied recalls of predicting Apps usage via different types of sensors for different users.}
	\label{fig:explicit}
\end{figure}

To show the prediction ability of different types of mobile sensors, we randomly select two users from the collected dataset and perform kNN classification via the three types of sensors respectively to predict their Apps usage. Figure~\ref{fig:explicit} shows the prediction recall of "Messenger", "Contacts", and "Browser" for the two users. As can be seen in Figure~\ref{fig:explicit}, personal sensors would be a good explicit feature for predicting $user_1$'s Apps usage, while environmental sensors are good for $user_2$. The reason is that $user_2$ probably needs a Wi-Fi signal to access the Internet.

\begin{table}
\centering
\caption{Hardware sensors for the explicit feature.}
	\begin{tabular}{ll}
	\hline
		Sensors & Contextual Information\\
	\hline\hline
		\multirow{4}*{Location} & Longitude\\
		& Latitude\\
		& Altitude\\
		& Location Cluster\\
	\hline	
		\multirow{2}*{Time} & Hour of day\\
		& Day of week\\
	\hline
		\multirow{2}*{Battery} & Battery Level\\
		& Charging status\\
	\hline
		\multirow{4}*{Accelerometer} & Avg. and std. dev. of \{x, y, z\}\\
		& Acceleration changes\\
		& speed\\
		& Heading\\
	\hline
		Wi-Fi Signal & Received signal\\
	\hline
		GSM Signal & Signal Strength\\
	\hline
		\multirow{2}*{System} & Free space of each drive\\
		& Free RAM \\
	\hline
	\end{tabular}
\label{tab:explicit}
\end{table}

\subsection{Implicit Feature Extraction}
The implicit feature formulates the usage transitions among Apps in a usage session. As mentioned in~\cite{DBLP:conf/mobisys/YanCGKL12}, users use a series of Apps, called a usage session, to complete a specific task. For example, one user could use "Maps" when travelling to a sightseeing spot, then use camera to take photos, and upload those photos to Facebook. Thus, the series of using "Maps", "Camera" and "Facebook" is called a usage session, denoted as "Map"$\xrightarrow{\delta_1}$"Camera"$\xrightarrow{\delta_2}$"Facebook", where $\delta_1$ and $\delta_2$ represent the transition intervals. 

The implicit feature of "Facebook" in this usage session is thus $<p_{MF}(\delta_1),p_{CF}(\delta_1 + \delta_2),p_{FF}(\infty)>$, where $p_{MF}(\cdot)$, $p_{CF}(\cdot)$, and $p_{FF}(\cdot)$ are probability models which represent the probability of using "Maps", "Camera" and "Facebook" respectively before using "Facebook" with the transition interval as the random variable. Note that because there is no "Facebook" to "Facebook" in this usage session, the transition interval is thus set to $\infty$ and then the probability would be 0. 

The probability model could be estimated from a user's historical usage trace. In this section, we introduce an Apps Usage Graph (AUG) which models the transition probability among Apps for a single user. For training purposes, the implicit features for the training usage sessions are derived by traversing the AUG. However, for testing purposes, since we do not know which is the App to be invoked, the derivation of the implicit feature for the training usage session cannot be utilized directly. Therefore, an iterative refinement algorithm is proposed to estimate both the next App and its implicit feature simultaneously. The following paragraphs will illustrate the details of the AUG construction and the implicit feature derivation for both the training and testing usage sessions.

\subsubsection{Apps Usage Graph (AUG)} 
\begin{figure}
\centering
\includegraphics[scale=0.8]{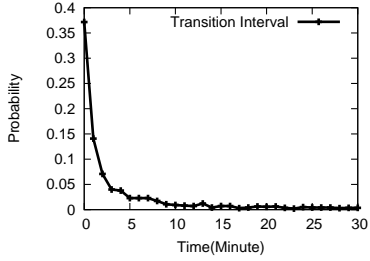}
\caption{The PDF of the duration of two consecutive App launches.}
\label{fig:transition}
\end{figure}

For each user, we construct an Apps Usage Graph (AUG) to describe the transition probability among Apps. An AUG is a directed graph where each node is an App, the direction of an edge between two nodes represents the usage order, and the weight on each edge is a probability distribution of the interval between two Apps. Since two consecutive launches could be viewed as a Poisson arrival process, we can formulate the intervals between two launches as an exponential distribution. For example, Figure~\ref{fig:transition} shows the probability density function (PDF) of two consecutive launches which exactly fulfils the exponential distribution where most transitions (e.g., 0.45\%) are within 1 minute.

Here, Equation~\ref{eq:exp} formulates the exponential density function of the launch interval being in $[x,x+1)$. The parameter $\alpha=\hat{p(0)}$ is derived by assigning $x=0$ in Equation~\ref{eq:exp}, and could be calculated by $p(0)$, the real probability derived from the training data. Then, $\beta$ is solved by minimizing the difference between the estimated probability $\hat{p(i)}$ and the real probability $p(i)$ as shown in Equation~\ref{eq:beta} for every interval $i$. 

Empirically, we do not need to fit every interval when obtaining the exponential model. For example, in Figure~\ref{fig:transition}, only the first 5 intervals already cover more than 75\% of the training data. Therefore, we can iteratively add one interval until the data coverage reaches a given threshold. We will discuss the impact of the data coverage threshold in the experiments section.

\begin{equation}
\hat{p(x)}=\alpha \exp^{-\beta x}
\label{eq:exp}
\end{equation}

\begin{eqnarray}
\beta &=&\argmin_{\beta} \sum_i|\hat{p(i)}-p(i)| \nonumber \\ 
      &=&\argmin_{\beta} \sum_i|p(0)\exp^{-\beta i}-p(i)|
\label{eq:beta}
\end{eqnarray}

\begin{figure}
\centering
\includegraphics[scale=0.45]{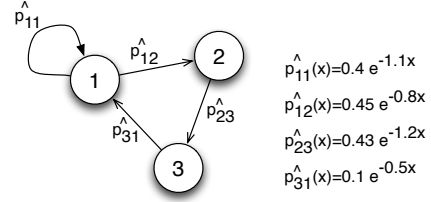}
\caption{An example of the Apps Usage Graph (AUG).}
\label{fig:graph}
\end{figure}

For example, Figure~\ref{fig:graph} shows an AUG with three Apps. From Figure~\ref{fig:graph}, the probability of two consecutive usages of $App_1$ with an interval of 0.3 minutes (i.e., $App_1 \xrightarrow{0.3} App_1$) is 0.4, and $App_1 \xrightarrow{1.5} App_2$ is 0.2. Although AUG only takes two consecutive Apps into account, such as $p_{12}$ and $p_{23}$, the probability of $p_{13}$, could be calculated by $p_{12} \times p_{23}$.

\subsubsection{Implicit Features for Training}
For each training case, the implicit features are derived by looking up the AUG. Suppose the currently used App (i.e., class label) is $App_t$, the implicit feature is thus, $<p'_{1t}, p'_{2t},...,p'_{nt}>$, where $p'_{it}$ represents the probability of transiting from $App_i$ to any random Apps and then to $App_t$. The probability of $p'^{(s)}_{it}$ is defined as in Equation~\ref{eq:training} which is the summation of every probability from $App_i$ to $App_t$. Note that we use a superscript, $s$, to indicate how many Apps are between $App_i$ and $App_t$, and $App_{m_k}$ is the $k$-th App after $App_i$. Once we derive the implicit feature in a reverse time order, the sub-problem of estimating $p'^{(s-k)}_{m_k,t}$ is already solved. The calculation of the implicit feature for $App_i$ stops when the transition probability falls below a given threshold, $min_{tp}$. In our collected dataset, the transition probability falls to 0.1\% when we look backward to more than 5 Apps, which is the default parameter for $min_{tp}$. Algorithm~\ref{algo:training} depicts the derivation of the implicit feature for a training case with $App_t$ as its class label. 

\begin{equation}
p'^{(s)}_{it}=\hat{p_{it}}+ \sum_k \hat{p_{i,m_k}} \times p'^{(s-k)}_{m_k,t}
\label{eq:training}
\end{equation}

\begin{algorithm}
	\KwIn{$App_t$: a training App}
	\KwOut{$IF_t$: the implicit feature of $App_t$}   
    	
	\BlankLine
		\ForEach{$App_i$ prior than $App_t$}{
			$IF_t[i] \leftarrow IF_t[i]+ \hat{p_{it}(\delta_{it})}$ \;				
			\ForEach{$App_m$ between $App_i$ and $App_t$}{
				$IF_t[i] \leftarrow IF_t[i] + \hat{p_{im}(\delta_{jm})} \times IF_m[t]$ \;				
			}			
		}		
    \Return $IF_t$
\caption{Deriving the implicit feature of $App_t$ for training.}
\label{algo:training}
\end{algorithm}

\begin{figure}
\centering
\includegraphics[scale=0.45]{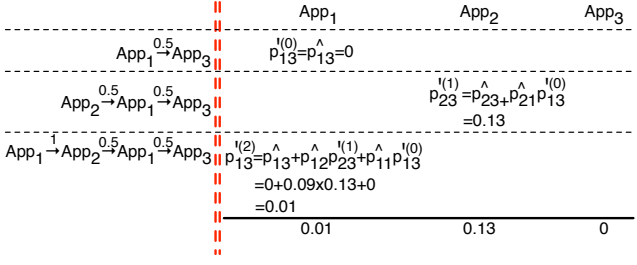}
\caption{Steps of obtaining the implicit feature of $App_3$ in the training case, $\dots \rightarrow App_1 \xrightarrow{1} App_2 \xrightarrow{0.5} App_1 \xrightarrow{0.5} App_3$.}
\label{fig:training}
\end{figure}

For example, suppose we have an AUG as shown in Figure~\ref{fig:graph} and a usage trace as $\dots \rightarrow App_1 \xrightarrow{1} App_2 \xrightarrow{0.5} App_1 \xrightarrow{0.5} App_3 \rightarrow \dots$. Figure~\ref{fig:training} shows the process of obtaining the implicit feature of $App_3$. We first estimate $p'^{(0)}_{13}$ from $App_1 \xrightarrow{0.5} App_3$, then $p'^{(1)}_{23}$ from $App_2 \xrightarrow{0.5} App_1 \xrightarrow{0.5} App_3$, and finally update $p'^{(2)}_{13}$ from $App1 \xrightarrow{1} App_2 \xrightarrow{0.5} App_1 \xrightarrow{0.5} App_3$. Note that $p'^{(0)}_{13}$ is reused for calculating $p'^{(1)}_{23}$, and $p'^{(1)}_{23}$ and $p'^{(0)}_{13}$ are reused for updating $p'^{(2)}_{13}$. The implicit feature of $App_3$ is $<0.01,0.13,0>$.

\subsubsection{Implicit Features for Testing}
Since the App to be predicted for current invocation, $App_t$, is unknown for testing, the derivation process of implicit features for training does not work. We propose an iterative refinement algorithm to estimate both $App_t$ and its implicit feature, $IF_t$, for testing. Suppose $\theta_i$ is the probability of $App_t=App_i$, the implicit feature $IF_t$ is calculated as in Equation~\ref{eq:if} which is a linear combination of the IF of each $App_i$. In addition, $M=[IF_1^T,IF_2^T,\dots]$ represents the transition matrix among Apps, where $IF_1^T$, $IF_2^T$, $\dots$ are column vectors. Then, the value of $\theta_i$ could be updated by Equation~\ref{eq:theta}, which is the probability of staying in $App_i$ after one-step walking along the transition matrix $M$. We keep updating $\theta_i$ and $IF_t$ iteratively, until $App_t$ is fixed to one specific App. In our experiments, the iterative refinement process converges in about 3 iterations. Algorithm~\ref{algo:testing} depicts the derivation of the implicit feature for testing.

\begin{equation}
IF_t=\sum\limits_{App_i} \theta_i \times IF_i
\label{eq:if}
\end{equation}

\begin{equation}
\theta_i=\sum\limits_{App_m} IF_t[m] \times M[m][i] 
\label{eq:theta}
\end{equation}

\begin{algorithm}
	\KwIn{$t$: a testing case}
	\KwOut{$IF_t$: the implicit feature at $t$}   
    	
	\BlankLine     
    	\While{$iter < threshold$} {
			\ForEach{$\theta_j$}{
				$IF_t \leftarrow IF_t + \theta_i \times IF_i$ \;
			}			
			\ForEach{$App_i$ prior than time $t$}{
				$\theta_i \leftarrow \theta_i + IF_t[m] \times M[m][i]$ \;
				Normalize $\theta_i$ \;			
			}    		
			$iter \leftarrow iter + 1$ \;						
		}
    \Return $IF_t$
\caption{Deriving the implicit feature for testing.}
\label{algo:testing}
\end{algorithm}

For example, suppose the testing case is $\dots \rightarrow App_1 \xrightarrow{1} App_2 \xrightarrow{0.5} App_1 \xrightarrow{0.5} App_t$. First, we initialize $\theta_i$ as $<1/3,1/3,1/3>$, which gives equal probability to each App, and the transition matrix $M=\left[\begin{array}{c c c}
0.49 & 0.6 & 0.01 \\
0 & 0 & 0.13 \\
0 & 0 & 0 \end{array} \right]$, which is derived by calculating the IF of each App shown in Equation~\ref{eq:training}. Note that the last row is all zero because there is no $App_3$ transiting to any other Apps. Then, the implicit feature is $<0.37,0.04,0>$ in the first iteration. Next, $\theta_i$ is updated to $<0.18,0.22,0.01>$, and normalized as $<0.44, 0.54, 0.02>$ according to one-step walk in $M$ with the calculated implicit feature as the prior probability. Then, we can obtain the implicit feature as $<0.53,0.01,0>$ in the second iteration.
\section{Personalized Feature Selection}
\label{sec:selection}
The goal of the personalized feature selection is to use as fewer features as possible to guarantee an acceptable accuracy. Due to the energy and storage consumption of collecting sensors readings and Apps transition relations, we should select useful features for different users in advance. Furthermore, through the personalized feature selection, we could avoid the curse of dimensionality on performing the kNN. We first apply the personalized feature selection on the training data, and then only the selected features are required to be collected in the future. 

Here, we propose a greedy algorithm to select the best feature iteratively. We adopt the concept of Minimum Description Length (MDL)~\cite{ris78,ris99} to evaluate the goodness of the features. For different features, we can have varied projections of the training data. We claim that if a feature needs fewer bits to describe its data distribution, it is good for predicting the data. Therefore, in each iteration, the feature with the minimum description length is selected. Then, those data points which are correctly predicted are logically eliminated from the training data, and the next feature is selected by the same process repeatedly. We define the description length of the hypothesis, which is shown in Equation~\ref{eq:lh}, as the length of representing the training data. $NG(App_i)$ is the number of groups of $App_i$. The description length of Data given the hypothesis is the total number of miss-classified data which is formulated as in Equation~\ref{eq:ldh}.

\begin{equation}
L(H)=\sum\limits_{i} \log_2 NG(App_i)
\label{eq:lh}
\end{equation}

\begin{equation}
L(D|H)=\sum\limits_{i} \log_2 (missClassified(App_i)+1)
\label{eq:ldh}
\end{equation}

\begin{figure}
\centering
\includegraphics[scale=0.35]{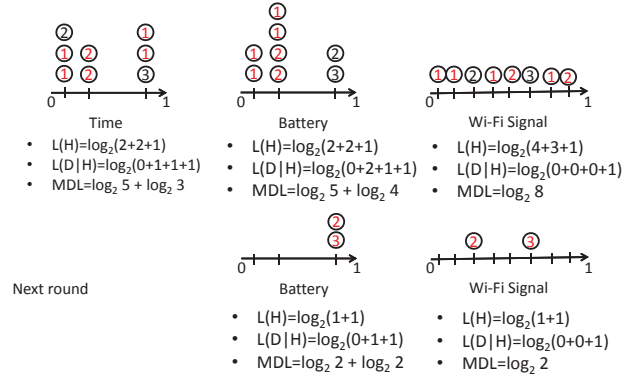}
\caption{An example of feature selection where the red data points are correctly predicted.}
\label{fig:mdl}
\end{figure}

For example, given 8 data points in the training data and three features as shown in Figure~\ref{fig:mdl}. In the first round, Time is the feature with minimum description length. Those data points marked as red are correctly predicted and will be removed. Therefore, in the second round, only two data points are left, and the feature of Wi-Fi signal is selected due to its minimum description length.

The selection process stops when a percentage of $\rho$ of the training data is covered. We also discuss the impact of $\rho$ in the experimental section. Note that the number of features affects the energy and storage consumption and is set according to the capability of the smartphones. Algorithm~\ref{algo:selection} depicts the process of personalized feature selection. After the selection, only the readings of the sensors which are selected will be collected as the explicit feature in the future. In addition, only the selected Apps will be used to construct AUG.   

\begin{algorithm}
	\KwIn{$D_z$: the training data}
	\KwOut{$PF$: the personalized features}   
    	
	\BlankLine
		Let $N_z\leftarrow|D_z|$ \;
		\While{$|D_z|< \rho N_z$}{
		\ForEach{feature $f$}{
			Calculate $DL_f$: description length for feature $f$ \;									
		}
		
		$PF \leftarrow PF \cup \{\argmax\limits_{f} DL_f\}$ \; 	
        Let $D_a$ be the set of accurately predicted data points \;		
		$D_z \leftarrow D_z-D_a$ \;
		}	
    \Return $PF$
\caption{Personalized feature selection.}
\label{algo:selection}
\end{algorithm}
\section{Experimental study}
\label{sec:exp}
In this section, we conduct a comprehensive set of experiments to compare the performance of the proposed KAP framework with other existing methods including 1) most frequently used (MFU) method, 2) most recently used (MRU) method which is the built-in prediction method in most mobile OS, such as Android and iOS, 3) SVM, 4) App Naive Bayes~\cite{DBLP:conf/huc/ShinHD12}, 5) Decision Tree, and 6) AdaBoost. In the following, we first discuss the collected dataset, then introduce the metrics employed to evaluate the performance, and finally deliver the experimental results.
 
\subsection{Dataset Description}
In this paper, we use a real world dataset collected by a mobile phone company which installed a monitoring program on every volunteer's smartphone. In this dataset, we have totally 50 volunteers including college students and faculty from June 2010 to January 2011. For each user, we separate the dataset into three parts, where each part consists of three months, and we use the first two months as training data, and the last one month as testing data. Totally, there are more than 300 different Apps installed on their smartphones, and the average number of Apps on one smartphone is 56. 

\subsection{Performance Metrics}
In this paper, we use two performance metrics: 1) average recall and 2) nDCG~\cite{DBLP:journals/tois/JarvelinK02} score.

\textbf{Average Recall:} Since there is only one App being launched in each testing case, recall score is thus adopted as one performance metric which evaluates whether the used App is in the prediction list. The recall score of one user is defined as $\sum\limits_{c_i \in C} \frac{I(App_{c_i}, L_{c_i})}{|C|}$, where $C$ is the set of testing cases, $App_{c_i}$ is the ground-truth, and $L{c_i}$ is the prediction list at the $i$-th testing case. $I(\cdot)$ is an indicator function which equals 1, when $App_{c_i} \in L_{c_i}$, and equals 0, otherwise. Finally, the average recall is the average of the recall values of all users.

\textbf{nDCG Score:} To evaluate the accuracy of the order of the prediction list, we also test the nDCG score of the prediction results. The IDCG score is fixed to 1 because there is only one used App in the ground-truth. The DCG score is $\frac{1}{\log_2(i+1)}$ when the used App is predicted at position $i$ of the prediction list. Then, nDCG is the average of $\frac{DCG}{IDCG}$ for all testing cases.

\subsection{Experimental Results}
To evaluate the performance of predicting Apps usage by the proposed KAP framework, we first evaluate the overall performance when predicting different numbers of Apps. Then, we test the performance of the personalized feature selection algorithm. The impact of different parameters for the KAP framework and kNN classification is also included. Note that we use top-$k=4$, kNN=40\%, and the minimum data coverage of personalized feature selection as 70\% to be the default parameter settings throughout the experiment.  

\subsubsection{Overall Performance}
\begin{figure}
	\centering
	\subfigure[Recall]{
    	\label{fig:sub:recall_k}
        \includegraphics[scale=0.5]{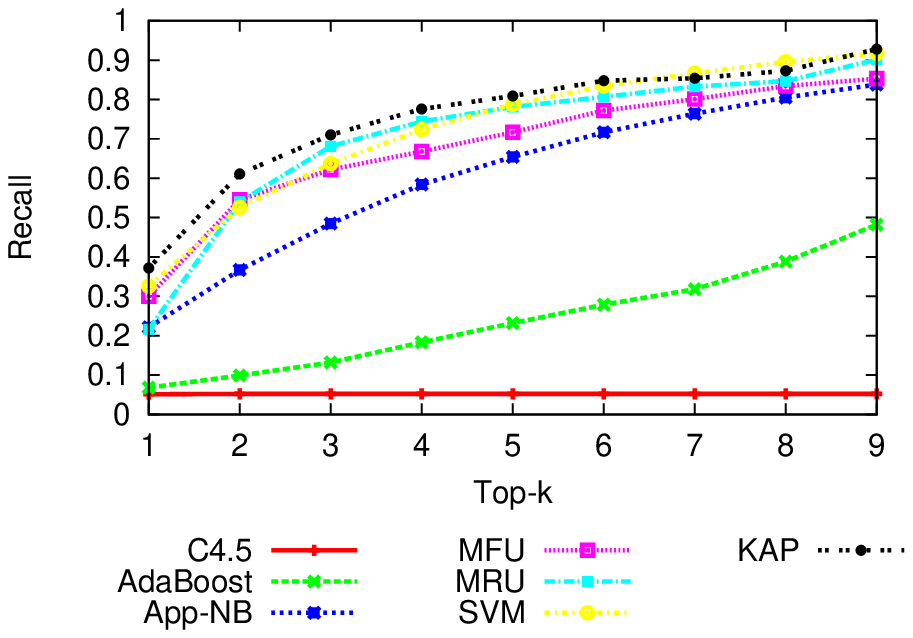}
    }
	\subfigure[nDCG]{
    	\label{fig:sub:ndcg_k}
        \includegraphics[scale=0.5]{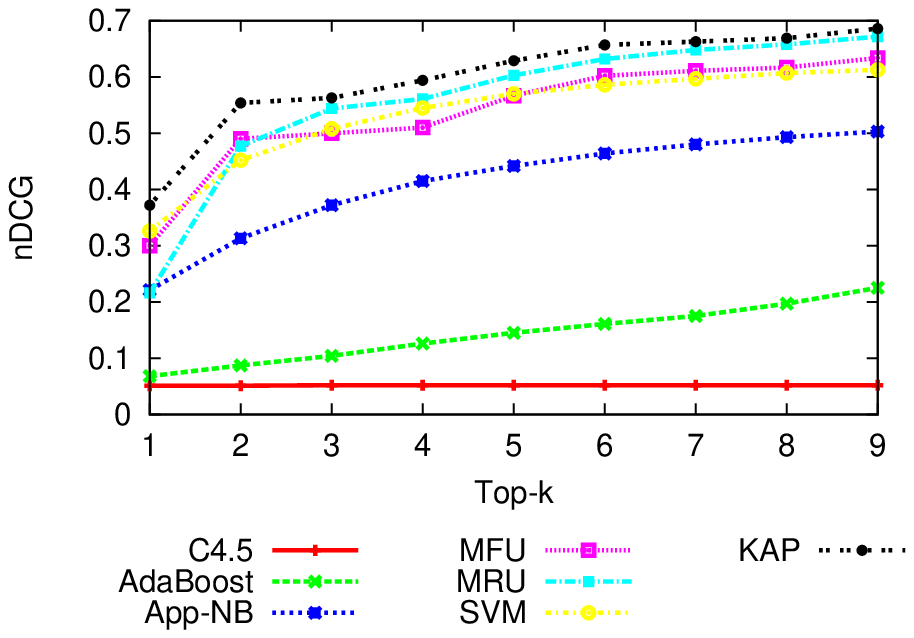}
    }
    \caption{Impact of the number of prediction, k.}
    \label{fig:k}
\end{figure}

First, we evaluate the performance KAP and other different methods under various numbers of prediction, $k$. As can be seen in Figure~\ref{fig:k}, when the number of prediction $k$ increases, both the recall and nDCG values also increase. However, KAP, MRU, MFU, and SVM perform better than others. In Figure~\ref{fig:sub:recall_k}, when $k=9$ (the number of predictions shown in the latest Android system), the recall of KAP could be more than 95\%, while it is only about 90\% for MFU, MRU, and SVM. On the other hand, the nDCG value of KAP shown in Figure~\ref{fig:sub:ndcg_k} is always higher than that of the other methods, which means the prediction order of KAP is better. 

\begin{figure}
	\centering
	\subfigure[Recall]{
    	\label{fig:sub:recall_f}
        \includegraphics[scale=0.5]{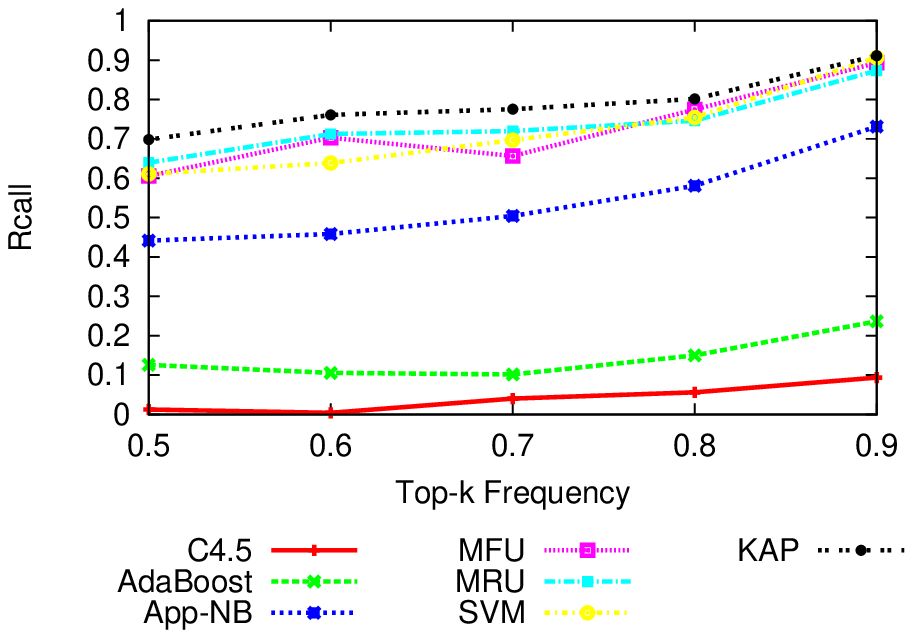}
    }
	\subfigure[nDCG]{
    	\label{fig:sub:ndcg_f}
        \includegraphics[scale=0.5]{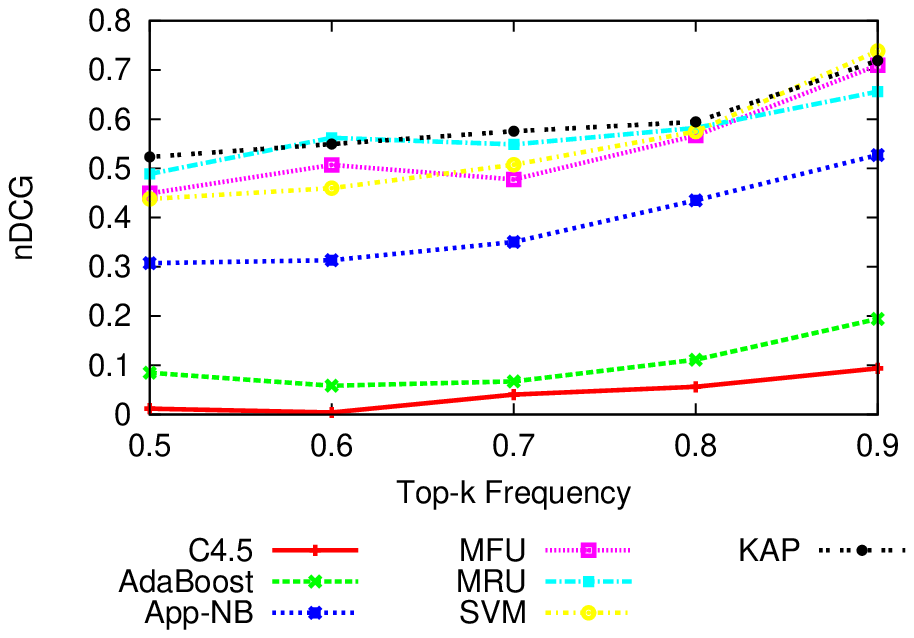}
    }
    \caption{Impact of top-k frequency.}
    \label{fig:f}
\end{figure}

Second, we test the accuracy of varied top-$k$ frequency. The top-$k$ frequency is defined as the ratio of the usage of the most frequent $k$ Apps. For example, if a user has 5 Apps and the usage counts are 3, 1, 2, 5, and 2, the top-$2$ frequency is thus $\frac{5+3}{3+1+2+5+2}=\frac{8}{13}$. Figure~\ref{fig:f} shows the results when top-$k=4$. Intuitively, when the top-$k$ frequency increases, the accuracy of the MFU method could be better. However, in Figure~\ref{fig:sub:recall_f}, even when the ratio is $0.9$, the MFU method performs just slightly better than the MRU method, but worse than both KAP and SVM. In Figure~\ref{fig:sub:ndcg_f}, the prediction order of KAP is also better than the results of the other methods. 

\subsubsection{Impact of Personalized Feature Selection}
For the proposed KAP method, we evaluate the performance of the personalized feature selection to see if the proposed MDL-based selection algorithm could reduce the used storage when maintaining a good prediction accuracy. For one user, the average used storage and prediction accuracy is shown in Table~\ref{tab:coverage} under different data coverage $\rho$. As can be seen in Table~\ref{tab:coverage} the personalized feature selection could reduce 55\% of training data size and only lose 1\% of recall and 3\% of nDCG when the data coverage is 70\%. In addition, Table~\ref{tab:time} compares the execution time of KAP with and without the personalized feature selection, where the training time is reduced dramatically under $\rho=70\%$.   

\begin{table}
\centering
\caption{The storage consumption and accuracy under varied data coverage $\rho$.}
	\begin{tabular}{lccccccccc}
	\hline
		Coverage(\%) & 30 & 40 & 50 & 60 & 70 & 80 & 90 & 100\\
	\hline\hline
		Storage(KB) & 28 & 31 & 34 & 37 & 43 & 52 & 82 & 94 \\
		Recall & 0.78 & 0.78 & 0.80 & 0.80 & 0.82 & 0.82 & 0.82 & 0.83 \\								
		nDCG & 0.50 & 0.51 & 0.52 & 0.53 & 0.55 & 0.57 & 0.57 & 0.58 \\
	\hline
	\end{tabular}
\label{tab:coverage}
\end{table}

\begin{table}
\centering
\caption{The execution time of KAP with and without personalized feature selection.}
	\begin{tabular}{lccc}
	\hline
		Execution time (ms) & Training & Testing & Total\\
	\hline\hline
		KAP & 86 & 160 & 246   \\								
		KAP without selection & 185 & 160 & 345 \\
	\hline
	\end{tabular}
\label{tab:time}
\end{table}

\subsection{Comparison of Different Usage Behavior}
Since different users have different usage behaivor, which could extremely affect the prediction accuracy. In this section, we separate users into different groups according to 1) number of installed Apps, 2) usage frequency, and 3) usage entropy. Then, we test the performance of applying different methods on different groups.

\subsubsection{Impact of the Number of Installed Apps}
\begin{figure}
	\centering
	\subfigure[Recall]{
    	\label{fig:sub:recall_apps}
        \includegraphics[scale=0.5]{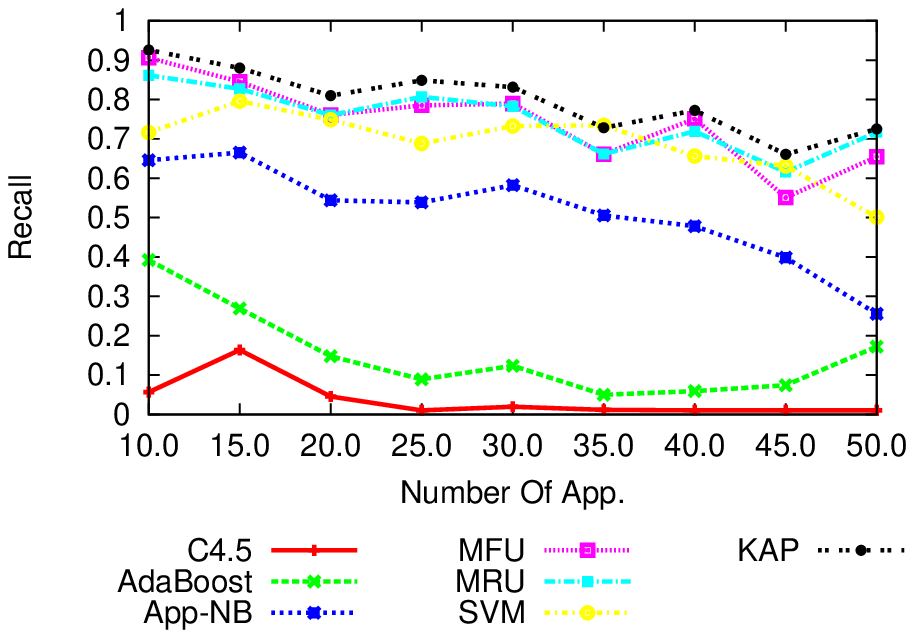}
    }
	\subfigure[nDCG]{
    	\label{fig:sub:ndcg_apps}
        \includegraphics[scale=0.5]{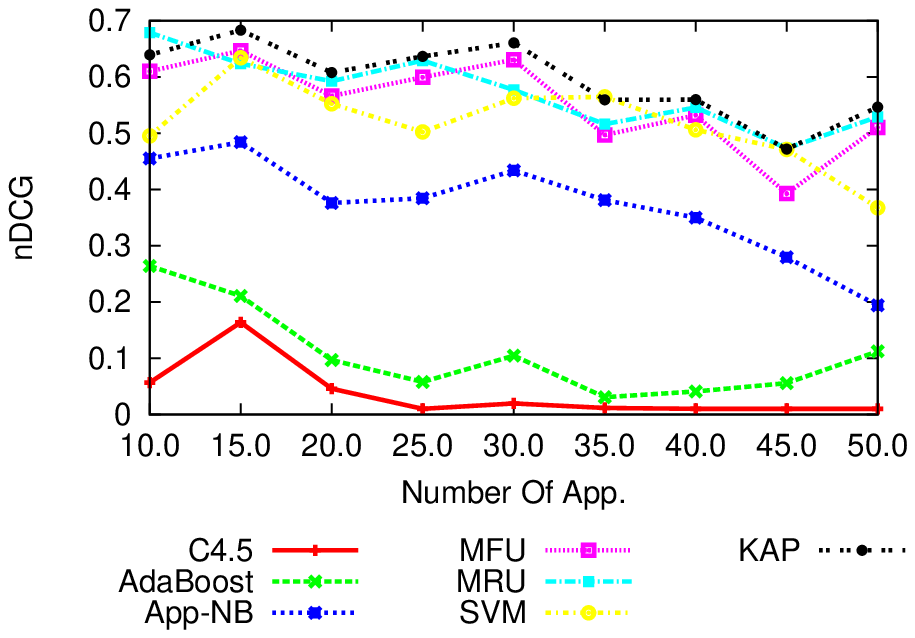}
    }
    \caption{Impact of the number of Apps.}
    \label{fig:apps}
\end{figure}

When users launch more Apps, it becomes more difficult to accurately predict Apps usage. Figure~\ref{fig:apps} shows the recall and ndcg results for a varying number of used Apps. As can be seen in Figure~\ref{fig:apps}, both the recall and ndcg values decrease when the number of used Apps increases for all methods. However, the decreasing rate of the proposed KAP method is much smoother than that of the others. The recall of KAP is around 85\% while that of the others is below 40\% when the number of used Apps is 30.

\begin{figure}
	\centering
	\subfigure[Recall]{
    	\label{fig:sub:recall_frequency}
        \includegraphics[scale=0.5]{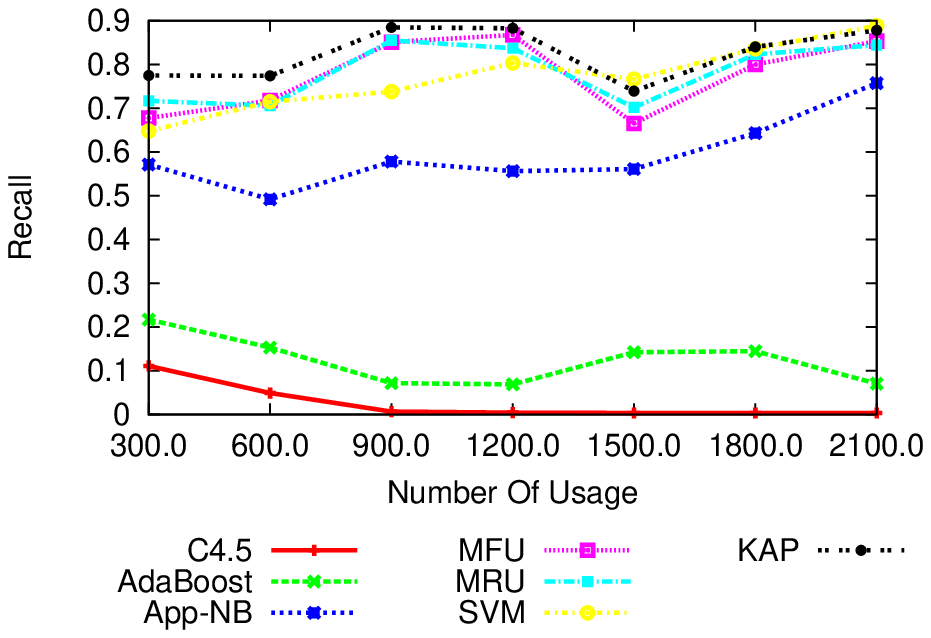}
    }
	\subfigure[nDCG]{
    	\label{fig:sub:ndcg_knn}
        \includegraphics[scale=0.5]{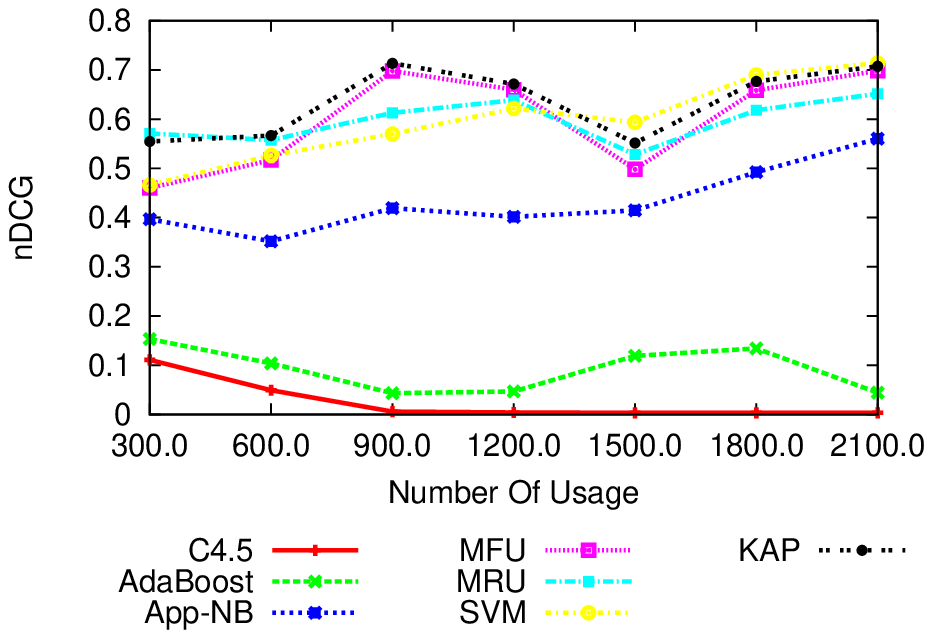}
    }
    \caption{Impact of the usage count.}
    \label{fig:frequency}
\end{figure}

\subsubsection{Impact of the Usage Count}
Now, we test the impact of the usage count. A higher usage count means we could have more training data to learn the classification model for App prediction. Concurrently, it provides more complicated information of users' usage behavior, and could make noisy data. Figure~\ref{fig:frequency} shows the recall and ndcg values. The performance of KAP, Naive Bayes, and SVM goes up when the usage count increases. However, AdaBoost and Decision Tree have worse performance as the usage count goes up. The result shows that the KAP algorithm can handle more complicated and noisy data.

\subsubsection{Impact of the Entropy of the Apps Usage}
\begin{figure}
	\centering
	\subfigure[Recall]{
    	\label{fig:sub:recall_entropy}
        \includegraphics[scale=0.5]{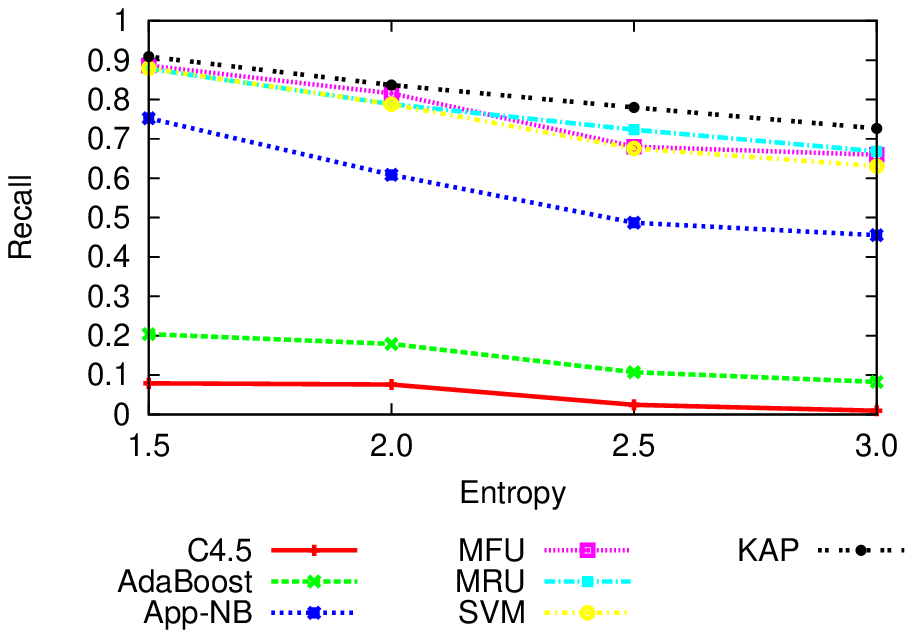}
    }
	\subfigure[nDCG]{
    	\label{fig:sub:ndcg_entropy}
        \includegraphics[scale=0.5]{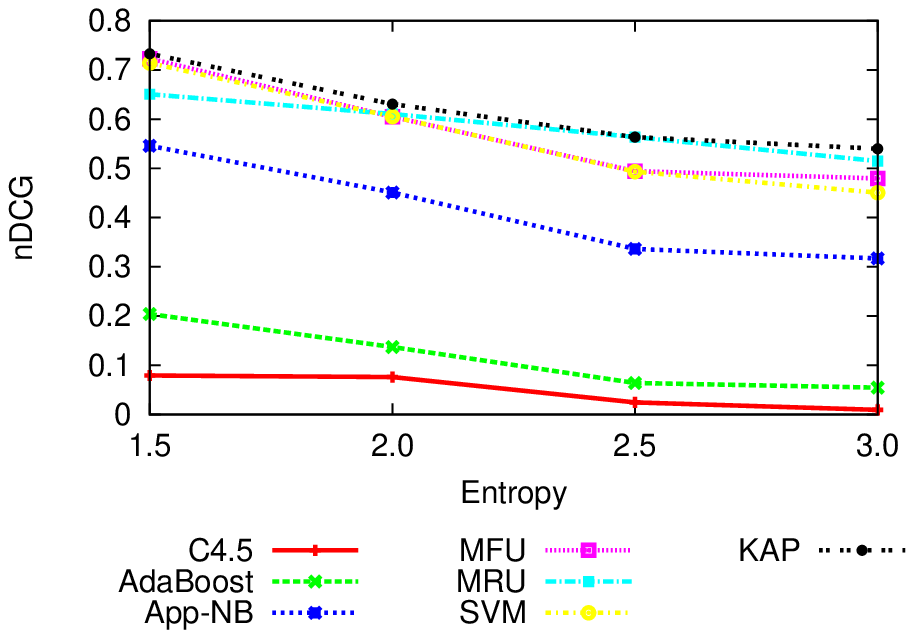}
    }
    \caption{Impact of the entropy of Apps.}
    \label{fig:entropy}
\end{figure}

We evaluate the impact of the entropy of the Apps usage. Intuitively, as the entropy of the Apps usage becomes larger, the Apps usage is almost random, and the performance of Apps usage prediction would become worse. Figure~\ref{fig:entropy} depicts that the proposed KAP could have around 50\% accuracy when the entropy goes to 3 where the other methods only have accuracy of less than 40\%.

\subsection{Impact of Different Parameters}
\subsubsection{Number of Iterations for Implicit Feature Extraction}
First, we test the number of iterations of deriving the implicit feature for each testing case. As shown in Table~\ref{tab:iteration}, the accuracy stays almost the same after the second iteration. This indicates that the iterative refinement algorithm could converge within 2 iteration which is sufficient to estimate the implicit feature.

\begin{table}
\centering
\caption{The recall and nDCG values under varied numbers of iterations.}
	\begin{tabular}{lccccc}
	\hline
		\#Iterations & 1 & 2 & 3 & 4 & 5\\
	\hline\hline
		Recall & 0.67 & 0.79 & 0.79 & 0.80 & 0.81  \\								
		nDCG & 0.43 & 0.59 & 0.59 & 0.60 & 0.61 \\
	\hline
	\end{tabular}
\label{tab:iteration}
\end{table}



\subsubsection{Minimum Probability for Identifying Usage Sessions}
As users usage sessions could be varied according to different tasks, we only need the useful length of the usage sessions to perform accurate Apps usage prediction, instead of calculate the full usage sessions. Therefore, we conduct this experiment to evaluate the impact of the length of usage sessions. Ac can be seen in Table~\ref{tab:session}, the results are not affected by the minimum transition probability, $min_{tp}$, too much. From our collected data, the session length is around 2 when $min_{tp}$ is 0.5, and the best case is under $min_{tp}=0.1$, which has the session length as around 5.

\begin{table}
\centering
\caption{The recall and nDCG values under varied minimum probability for session identification.}
	\begin{tabular}{lccccccc}
	\hline
		$min_{tp}$ & 0.5 & 0.25 & 0.1 & 0.075 & 0.05 & 0.025 & 0.001\\
	\hline\hline
		Recall & 0.73 & 0.77 & 0.83 & 0.81 & 0.80 & 0.75 & 0.74  \\								
		nDCG & 0.53 & 0.57 & 0.61 & 0.58 & 0.55 & 0.53 & 0.52 \\
	\hline
	\end{tabular}
\label{tab:session}
\end{table}

\subsubsection{Parameters for kNN Classification}
Finally, we evaluate the impact of selecting different numbers of neighbors to perform kNN classification. Here, we fix the number of predictions to 4 Apps and compare the recall and nDCG values of KAP and the other methods. Because the training data of different users could vary from several hundreds to thousands. we use a relative value for the number of neighbors. Table~\ref{tab:knn} shows the results of the recall and nDCG values for different number of neighbors. As can be seen in Table~\ref{tab:knn}, even only select 40\% of training data as the neighbors, the recall value is almost 80\%. Therefore, we set the default number of neighbor as 40\% throughout the experiments.

\begin{table}
\centering
\caption{The recall and nDCG values under varied number of neighbors for kNN.}
	\begin{tabular}{lccccc}
	\hline
		kNN(\%) & 20 & 40 & 60 & 80 & 100\\
	\hline\hline
		Recall & 0.74 & 0.79 & 0.80 & 0.80 & 0.81 \\								
		nDCG & 0.55 & 0.61 & 0.63 & 0.63 & 0.64 \\
	\hline
	\end{tabular}
\label{tab:knn}
\end{table}

\section{conclusion}
\label{sec:conclusion}
In this paper, we propose an Apps usage prediction framework, KAP, which predicts Apps usage regarding both the explicit readings of mobile sensors and the implicit transition relation among Apps. For the explicit feature, we consider three different types of mobile sensors: 1) device sensors, 2) environmental sensors, and 3) personal sensors.  For the implicit features, we construct an Apps Usage Graph (AUG) to model the transition probability among Apps. Then, for each training datum, we could represent the next used App as the implicit feature which describes the probability of transition from other Apps. Note that, since the next App in the testing data is unknown, we propose an iterative refinement algorithm to estimate both the probability of the App to be invoked next and its implicit feature.
We claim that different usage behaviors are correlated to different types of features. Therefore, a personalized feature selection algorithm is proposed, where for each user, only the most relative features are selected. Through the feature selection, we can reduce the dimensionality of the feature space and the energy/storage consumption. 

We integrate the explicit and implicit features as the feature space and the next used App as the class label to perform kNN classification. In the experimental results, our method outperforms the state-or-the-art methods and the currently used methods in most mobile devices. In addition, the proposed personalized feature selection algorithm could maintain better performance than using all features. We also evaluate the performance of KAP for different types of users, and the results show that KAP is both adaptive and flexible.



\bibliographystyle{IEEEtran}
\bibliography{icdm}
%



\end{document}